\pdfoutput=1

\documentclass[11pt]{article}

\usepackage{emnlp2021}

\usepackage{times}
\usepackage{latexsym}
\usepackage{graphicx}
\usepackage{amsmath}
\usepackage{upquote}
\usepackage{bm}
\usepackage{booktabs}
\usepackage{subcaption}

\usepackage[T1]{fontenc}

\usepackage[utf8]{inputenc}

\usepackage{microtype}

\title{Empathetic Dialog Generation with Fine-Grained Intents}

\author{Yubo Xie \and Pearl Pu \\
  School of Computer and Communication Sciences \\
  \'{E}cole Polytechnique F\'{e}d\'{e}rale de Lausanne, Switzerland \\
  \texttt{\{yubo.xie,\,pearl.pu\}@epfl.ch} \\}

\begin{document}
\maketitle
\begin{abstract}
Empathetic dialog generation aims at generating coherent responses following previous dialog turns and, more importantly, showing a sense of caring and a desire to help. Existing models either rely on pre-defined emotion labels to guide the response generation, or use deterministic rules to decide the emotion of the response. With the advent of advanced language models, it is possible to learn subtle interactions directly from the dataset, providing that the emotion categories offer sufficient nuances and other non-emotional but emotional regulating intents are included. In this paper, we describe how to incorporate a taxonomy of 32 emotion categories and 8 additional emotion regulating intents to succeed the task of empathetic response generation. To facilitate the training, we also curated a large-scale emotional dialog dataset from movie subtitles. Through a carefully designed crowdsourcing experiment, we evaluated and demonstrated how our model produces more empathetic dialogs compared with its baselines.
\end{abstract}

\section{Introduction}
\begin{figure}[t]
    \centering
    \includegraphics[width=\columnwidth]{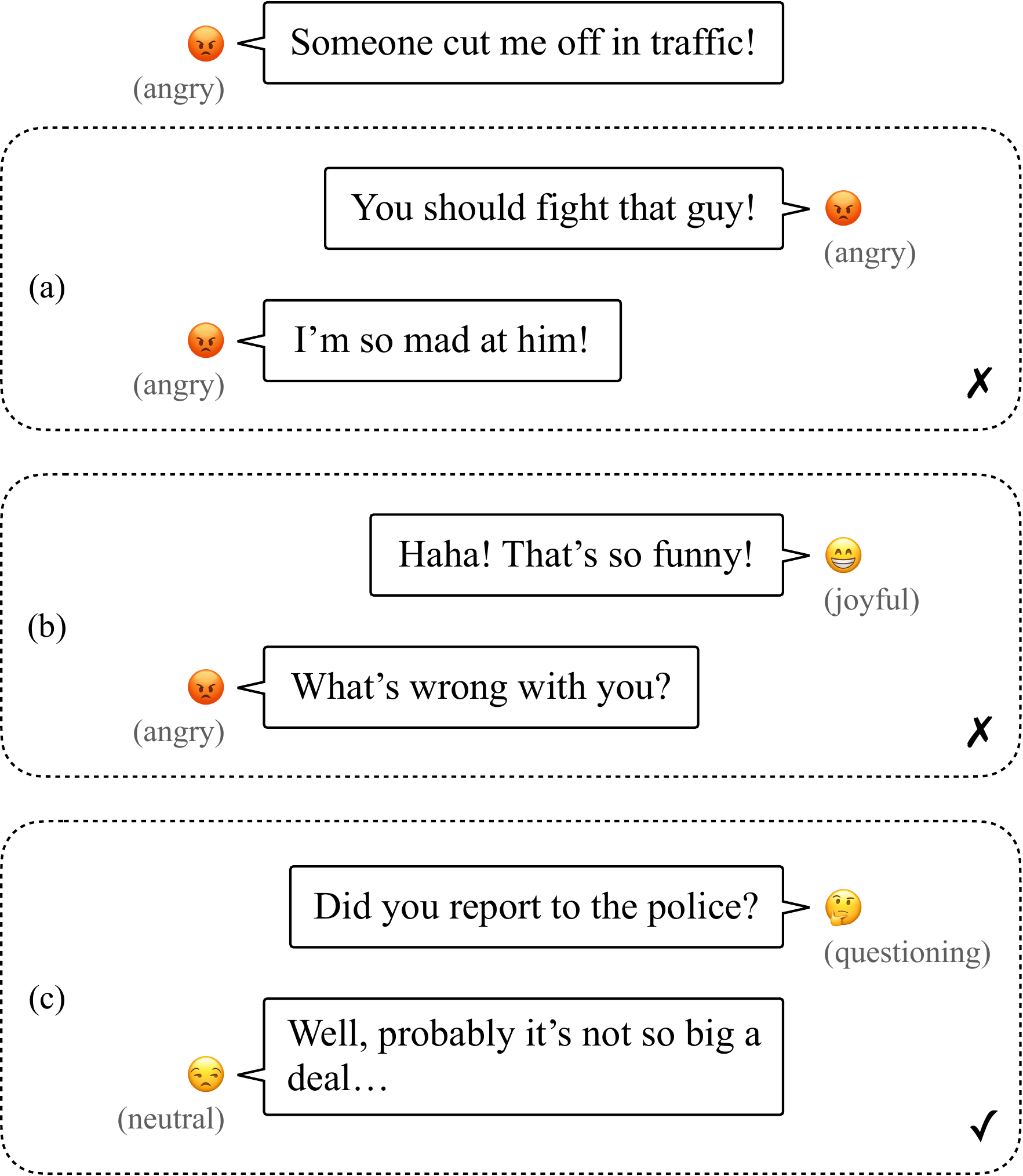}
    \caption{Three ways of responding to a speaker's utterance. Note that simply following the speaker's emotion state (a) or reversing it (b) still leaves the speaker in angry state (or even escalates the situation). Responding with questioning (c) successfully calms down the speaker and drives the conversation to a more manageable direction.}
    \label{fig:conv_example}
\end{figure}
Empathy is considered to be an innate ability of human beings~\cite{roth2011empathy} and plays an important role in people's social communication~\cite{valente2016empathy}. It has been shown that integrating empathy into dialog systems could improve user experience for human-computer interaction~\cite{liu2005embedded}. One of the empathetic components is the capacity to respond with an appropriate emotion to another person's mental states~\cite{shamay2009two}. In this regard, many existing neural dialog systems~\cite{DBLP:conf/aaai/ZhouHZZL18,DBLP:conf/naacl/HuangZTD18,DBLP:conf/acl/WangZ18,DBLP:conf/naacl/ColomboWMKK19,DBLP:conf/acl/SongZLXH19,DBLP:conf/acl/ShenF20} generate emotional responses conditioned on a pre-specified emotion label. However, this might be impractical when deploying the chatbots in reality, since an extra label is required as input. Other neural dialog systems~\cite{DBLP:conf/ecir/AsgharPHJM18,DBLP:conf/emnlp/LiS18,DBLP:conf/aaai/Zhong0M19,DBLP:conf/emnlp/LinMSXF19,DBLP:journals/corr/abs-2009-09708} adopt manually defined rules, either explicitly or implicitly, to decide the emotion state for the response to be generated, e.g., following/reversing the speaker's emotion, or just maximizing the emotion content in the response. However, such deterministic rules are not confirmed by psychology literature, and they ignore the subtle interactions captured in human conversations, where the listener often exhibits empathetic intents that are more neutral. Figure~\ref{fig:conv_example} gives an example of a situation where responding with the same or opposite emotion fails to drive the conversation towards an empathetic direction. In fact, as revealed by \citet{DBLP:conf/coling/WelivitaP20}, listeners are much more likely to respond with \emph{questioning} to sad or angry emotions of another person, than expressing similar or opposite emotions.

Therefore, it is necessary to incorporate these additional empathetic response intents explicitly into the design of dialog systems. Existing neural dialog systems adopt an empathetic dialog dataset that either has no neutral category~\cite{DBLP:conf/acl/RashkinSLB19}, or the neutral category is a conglomerate of intents that cannot be clearly defined. This is why this category is often called \emph{other}, which shows it is not sufficiently treated~\cite{DBLP:conf/semeval/ChatterjeeNJA19,DBLP:conf/ijcnlp/LiSSLCN17}. While \citet{DBLP:journals/corr/abs-1807-07255} proposed to model open-domain dialog generation as the selection of dialog acts that control the generation of responses, they did not specifically focus on the generation of empathetic dialogs.

Based on the taxonomy proposed by \citet{DBLP:conf/coling/WelivitaP20}, we incorporated an extra set of eight empathetic response intents (\emph{questioning}, \emph{agreeing}, \emph{acknowledging}, \emph{sympathizing}, \emph{encouraging}, \emph{consoling}, \emph{suggesting}, and \emph{wishing}) plus \emph{neutral} into the design of an empathetic dialog model, in addition to the 32 emotion categories proposed by \citet{DBLP:conf/acl/RashkinSLB19}. Emotional experience is primarily a reaction to an external event, for example, a loud sound, a surprising result on an exam, etc. In the case of dialogs, this emotional experience is shared by the interlocutors. When a listener wants to acknowledge or console the speaker, for example, he or she is expressing an emotional intent. This is the reason we treat all of these additional categories as well as the 32 emotion categories as dialog intents.

Overall speaking, our contributions are as follows: (1) We are the first to consider modeling a fine-grained set of empathetic response intents in an empathetic dialog model, which ensures a more precise learning of the emotional interactions revealed in the dialog data; (2) To facilitate the training of our empathetic dialog model, we curated a large-scale dialog dataset from movie subtitles; (3) To effectively evaluate our empathetic dialog model, we carefully designed a crowdsourcing experiment that enabled the workers to work on the tasks more easily. A total number of 6,000 dialogs were evaluated, which, to our knowledge, has never been attempted before for the evaluation of empathetic dialog systems.

\section{Related Work}
\paragraph{Neural dialog generation}
Dialog generation has been treated as a sequence transduction problem since the advent of deep neural models. \citet{DBLP:journals/corr/VinyalsL15} trained the seq2seq network on IT helpdesk dialogs and OpenSubtitles data. \citet{DBLP:conf/acl/ShangLL15} applied an attention mechanism to the seq2seq network and trained it on short-text social media dialogs. To adapt the seq2seq model to a multi-turn setting, \citet{DBLP:conf/aaai/SerbanSBCP16} designed a hierarchical encoder-decoder structure, based on which \citet{DBLP:conf/aaai/XingWWHZ18} devised a hierarchical attention mechanism so that the model could pay attention at both token-level and utterance-level.

\paragraph{Empathetic dialog generation}
\citet{DBLP:conf/aaai/LubisSYN18} designed a hierarchical encoder-decoder model that captures the user's emotion state and takes it into account when generating the response. \citet{DBLP:journals/corr/abs-1908-07816} proposed a multi-turn emotionally engaging dialog model by modeling the emotion states in the dialog history. \citet{DBLP:conf/icassp/ShinXMF20} adopted a reinforcement learning framework that provides a higher reward to the generative model if it promotes the user's future emotion state. \citet{DBLP:journals/corr/abs-1911-08698} adopted an adversarial learning framework and proposed two discriminators to evaluate if the generated response is empathetic and elicits more positive emotions by considering the emotion words in the gold response and the next reply.

\paragraph{Emotional dialog datasets}
Most of the existing emotional dialog datasets are small in size and have limited number of emotion categories. \citet{DBLP:conf/ijcnlp/LiSSLCN17} created the DailyDialog dataset from English learning websites, consisting of 13K multi-turn dialogs manually labeled with 7 emotions. The EmotionLines dataset~\cite{DBLP:conf/lrec/HsuCKHK18} contains 2,000 dialogs collected from Friends TV scripts and EmotionPush chat logs, labeled with 7 emotions. \citet{DBLP:conf/acl/PoriaHMNCM19} extended the EmotionLines dataset to a multimodal setting, containing 1,433 dialogs from Friends TV scripts. \citet{DBLP:conf/semeval/ChatterjeeNJA19} proposed the EmoContext dataset collected from users' interaction with a conversational agent, which contains 38K dialogs labeled with 4 emotions. \citet{DBLP:conf/acl/RashkinSLB19} curated the EmpatheticDialogues dataset containing 25K dialogs collected from a crowdsourcing platform by letting workers communicate with each other based on 32 emotion categories.

\section{Data Curation}

Existing empathetic dialog corpora are usually limited in size and training solely on these datasets could not give us a chatbot with desirable performance. Therefore, we would like to take advantage of transfer learning and pre-train the dialog model on a huge amount of dialog data (not necessarily empathetic), and then fine-tune it on a possibly much smaller empathetic dialog dataset.

\subsection{Extracting Dialogs from Movie Subtitles}
\label{subsec:os_dialogs}

\begin{table}[t]
    \centering
    \begin{tabular}{ll}
        \toprule
        Total \# dialogs & 4,010,009 \\
        Total \# turns & 18,849,440 \\
        Total \# tokens & 312,574,468 \\
        Average \# turns per dialog & 4.70 \\
        Average \# tokens per turn & 16.58 \\
        Average \# tokens per dialog & 77.95 \\
        \bottomrule
    \end{tabular}
    \caption{OpenSubtitles dialogs after cleaning.}
    \label{tab:os_dialogs}
\end{table}

To obtain a large-scale dialog dataset, we relied on the OpenSubtitles2018 corpus~\cite{lison2019open}, which contains text collected from movie subtitles spread over 60 languages, and is a good source of human conversations written by professional screenwriters. We only used the English part, which has 447K subtitle files, 441M sentences and 3.2B tokens. Due to the lack of speaker information in the OpenSubtitles corpus, before extracting the dialogs, we followed the same procedure proposed by \citet{DBLP:conf/slt/LisonM16} and built an SVM classifier to determine whether two consecutive lines in one subtitle file are actually spoken by the same character and should be in the same dialog turn. As a result, we obtained a turn segmentation accuracy of 76.69\%.

We then separated these turns into dialogs by adopting a heuristic rule based on timestamps: for each subtitle file, we calculate the gap between the starting time of each turn and the ending time of its previous turn. If this time gap is greater than 5 seconds, we cut off at this position and regard these two turns as belonging to different dialogs. An exception is when the timestamp information is missing for one of the two turns. In this case, we just regard them as belonging to one dialog. In this way, we obtained 9M dialogs from the whole English OpenSubtitles corpus. To further clean the dataset, we applied a sequence of steps to remove undesirable utterances. As a result, we obtained 4M cleaned OpenSubtitles dialogs. See Appendix~\ref{appendix:cleaning} for the detailed cleaning procedure. Table~\ref{tab:os_dialogs} lists some statistics of the OpenSubtitles dialogs.

\subsection{Emotional Dialogs in OpenSubtitles}
\label{subsec:emotional_os_dialogs}

\begin{table}[t]
    \centering
    \begin{tabular}{ll}
        \toprule
        Total \# dialogs & 1,000,000 \\
        Total \# turns & 3,488,300 \\
        Total \# tokens & 66,447,274 \\
        Average \# turns per dialog & 3.49 \\
        Average \# tokens per turn & 19.05 \\
        Average \# tokens per dialog & 66.45 \\
        \bottomrule
    \end{tabular}
    \caption{Emotional Dialogs in OpenSubtitles.}
    \label{tab:edos_dialogs}
\end{table}

\begin{figure*}[t]
    \centering
    \includegraphics[width=\textwidth]{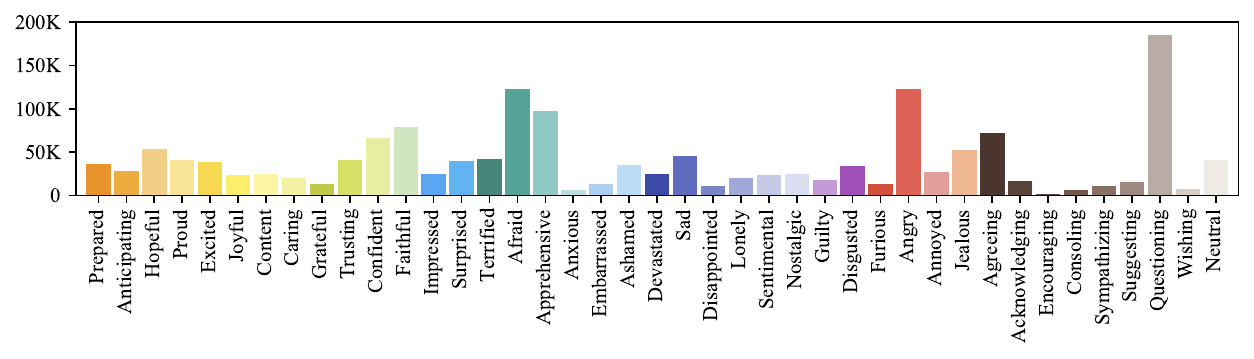}
    \caption{Distribution of emotions/intents in the emotional dialogs in OpenSubtitles.}
    \label{fig:edos_emot_dist}
\end{figure*}

Many existing emotional dialog datasets are small in size due to the expensive procedure of data collection, usually done manually by human. In this paper, we created a large-scale empathetic dialog dataset by first training a sentence-level fine-grained emotion classifier and then selecting out emotional dialogs from the cleaned OpenSubtitles dataset aforementioned.

To build the emotion classifier, we followed \citet{DBLP:conf/coling/WelivitaP20} and fine-tuned RoBERTa~\cite{DBLP:journals/corr/abs-1907-11692} on the situation sentences from the EmpatheticDialogues~\cite{DBLP:conf/acl/RashkinSLB19} training set (labeled with 32 fine-grained emotions), and 7K listener utterances labeled with 8 empathetic intents (\emph{questioning}, \emph{agreeing}, \emph{acknowledging}, \emph{sympathizing}, \emph{encouraging}, \emph{consoling}, \emph{suggesting}, and \emph{wishing}) plus one \emph{neutral} category (all other not mentioned intents). The 7K intent-labeled utterances were obtained by first manually labeling 521 sentences and then expanding through searching most frequent $n$-grams for each intent. The classifier achieved an accuracy of 65.88\% on the EmpatheticDialogues test set. We applied the obtained classifier on all cleaned OpenSubtitles dialogs, and calculated a probability distribution over the 41 categories for each utterance. We then define the emotionality of each utterance as the sum of the probability values of the 32 emotion categories, and the emotionality of each dialog as the averaged emotionality values of its utterances. We selected the top 1M dialogs with highest emotionality values to form the dataset of emotional dialogs in OpenSubtitles. Table~\ref{tab:edos_dialogs} lists the statistics of this dataset, and Figure~\ref{fig:edos_emot_dist} gives the distribution of emotions/intents of the last utterance. Some samples of the OpenSubtitles dialogs can be found in Appendix~\ref{appendix:more_samples}. The datasets along with the code of our model are publicly available.\footnote{\url{https://github.com/yuboxie/meed2}}

\section{An Empathetic Dialog Model}

\begin{figure*}[t]
    \centering
    \begin{subfigure}[t]{0.475\textwidth}
        \centering
        \includegraphics{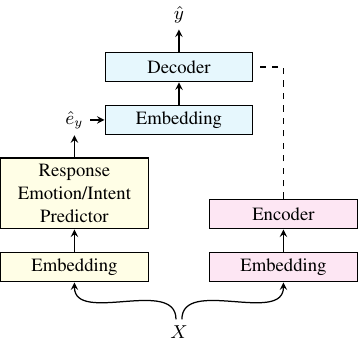}
        \caption{Overall architecture showing how the model works in inference mode. Dashed line denotes multi-head attention.}
        \label{fig:overall_model}
    \end{subfigure}
    \quad
    \begin{subfigure}[t]{0.475\textwidth}
        \centering
        \includegraphics{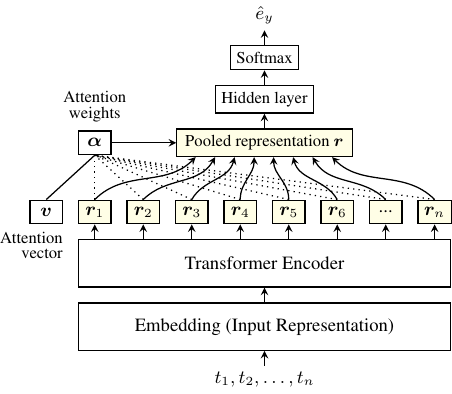}
        \caption{A detailed illustration of the response emotion/intent predictor. Dotted lines denote attention mechanism.}
        \label{fig:response_emot_pred}
    \end{subfigure}
    \caption{Illustrations of our dialog model.}
    \label{fig:model}
\end{figure*}

We propose an empathetic dialog model that incorporates the fine-grained set of empathetic response intents, by training a classifier that predicts the response emotion/intent, and based on that, generates the response accordingly. Compared with our previous multi-turn emotionally engaging dialog model (MEED)~\cite{DBLP:journals/corr/abs-1908-07816}, this model has more fine-grained emotions/intents, and allows for more controllability over the generated responses.

The problem could be defined as follows: given a dialog context $X$ consisting of one or more utterances $u_1,u_2,\dots,u_m$, spoken between two people, try to generate a response $\hat{y}$ that not only follows the dialog context but also is emotionally appropriate. Our model consists of three modules: (1) an encoder responsible for encoding the input $X$ into vector representations; (2) a response emotion/intent predictor which takes $X$ as input and decides in which emotion/intent the model should respond; (3) a decoder responsible for generating the actual response. We use Transformer~\cite{DBLP:conf/nips/VaswaniSPUJGKP17} encoder structure for our encoder and emotion/intent predictor, and Transformer decoder structure for our decoder. Figure~\ref{fig:overall_model} gives an overall depiction of the whole model architecture. All the three modules have the same input representation, which we describe in detail next.

\subsection{Input Representation}

\begin{figure*}[t]
    \centering
    \includegraphics{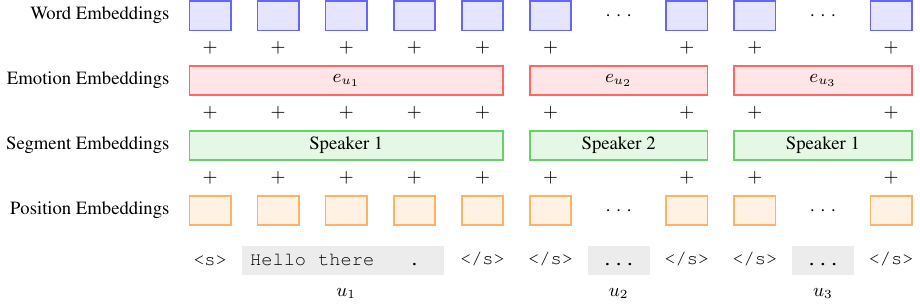}
    \caption{Input representation of our dialog model.}
    \label{fig:input_rep}
\end{figure*}

The input representation is illustrated in Figure~\ref{fig:input_rep}. We use the RoBERTa tokenizer to tokenize the utterances $u_1,u_2,\dots,u_m$ in the input dialog context $X$, and concatenate them by two special tokens: \texttt{<s>} and \texttt{</s>}, as shown in the figure. For our model to have a better understanding of the input dialog context, in addition to the word embeddings and position embeddings in the original Transformer architecture, we also have emotion embeddings. Specifically, for each utterance $u_i$, we use the same emotion classifier described in Section~\ref{subsec:emotional_os_dialogs} to obtain an emotion representation in the form of a probability distribution on 41 emotions/intents. The label with maximum probability value is denoted as $e_{u_i}$, representing the emotion/intent expressed by utterance $u_i$. Similar to word embeddings, we embed this emotion/intent $e_{u_i}$ into a vector space with the same dimensionality as other embeddings, so that they could add up. The same emotion embedding is used for all the tokens in the same utterance. To further differentiate between the speakers, we augment the input representation with segment embeddings. Utterances spoken by the same person would have the same segment embedding. The encoder and decoder share the same embedding tables.

\subsection{Response Emotion/Intent Predictor}
We relied on a data-driven approach to decide the emotion/intent of the response to be generated, by designing an emotion/intent classifier to predict the emotion/intent of the ground-truth response $y$, based on the context $X$. As shown in Figure~\ref{fig:response_emot_pred}, we use a Transformer encoder to get a context-dependent vector representation $\bm{r}_i$ for each of the input token $t_i$. To pool these high-level representations into a single vector, we use a simple attention mechanism and incorporate a trainable vector $\bm{v}$ to obtain an attention weight $\bm{\alpha}_i$ for $\bm{r}_i$,
\begin{equation}
    \bm{\alpha}_i = \frac{\exp(\bm{v}^T\bm{r}_i)}{\sum_{j=1}^{N}\exp(\bm{v}^T\bm{r}_j)}.
\end{equation}
The aggregate representation $\bm{r}$ is then
\begin{equation}
    \bm{r} = \sum_{i=1}^N \bm{\alpha}_i \bm{r}_i.
\end{equation}
$\bm{r}$ is fed into a hidden layer followed by a softmax layer to produce $\hat{e}_y$, denoting the predicted emotion/intent of the response to be generated.

\subsection{Training}
The response emotion/intent predictor is trained separately from the encoder/decoder, which means the training phase is a bit different from what is illustrated in Figure~\ref{fig:overall_model}. In particular, the response emotion/intent predictor is independently trained to minimize the cross entropy loss of $\hat{e}_y$ with respect to $e_y$ (true emotion/intent of $y$). While training the encoder and decoder simultaneously, we just feed $e_y$ into the embedding layers of the decoder, and try to minimize the cross entropy loss of $\hat{y}$ with respect to ${y}$.

We also experimented with jointly training the response emotion/intent predictor and the encoder/decoder, by combining two loss functions like in a multi-task setting. However, we found the generated responses quite generic compared with training the two components separately, plus joint training also introduces more hyperparameters to be tuned. Moreover, having them trained separately endows the decoder with more controllability---the decoder is able to generate responses according to a specified emotion/intent label.

\section{Evaluation}

We trained our empathetic dialog model and the baselines on three datasets and evaluated them in \emph{held-out} setting (meaning the test data comes from the same domain as the training data) and \emph{zero-shot} setting (meaning the test data comes from a different domain than the training data), using both automatic metrics and human judgement via crowdsourcing.

\subsection{Datasets}
Three datasets were involved in the evaluation:
\begin{itemize}
    \item \textbf{OpenSubtitles dialogs}. As described in Section~\ref{subsec:os_dialogs}, these dialogs were obtained by segmenting the movie subtitles. Note that for the purpose of pre-training, we excluded the emotional dialogs in OpenSubtitles (containing 1M dialogs), resulting in around 3M dialogs. We denote this dataset as OS.
    \item \textbf{Emotional dialogs in OpenSubtitles}. The curation process is described in Section~\ref{subsec:emotional_os_dialogs}. The total number of dialogs is 1M. We denote this dataset as EDOS.
    \item \textbf{EmpatheticDialogues dataset}. This dataset is created by~\citet{DBLP:conf/acl/RashkinSLB19} and contains 24,850 dialogs collected from crowdsourcing. We denote this dataset as ED.
\end{itemize}
We split each of the three datasets into training set (80\%), validation set (10\%), and test set (10\%). Among the dialogs of each test set, we further randomly selected out 2,000 to form a combined test set of 6,000 dialogs, for the purpose of evaluating the models on automatic metrics and human judgement via crowdsourcing.

\subsection{Baselines}
Similar to the work of~\citet{DBLP:conf/acl/RashkinSLB19}, we adopted the full Transformer model as our baseline, and based on the training strategies, we have the following variants:
\begin{itemize}
    \item \textbf{Pre-trained}. To take advantage of transfer learning, we pre-trained the full Transformer model on the curated OS dataset, which contains around 3M dialogs. The large scale of this training set is expected to provide a good starting point for fine-tuning.
    \item \textbf{Fine-tuned}. We took the pre-trained full Transformer, and then fine-tuned it on two smaller dialog datasets: our curated EDOS dataset, and the ED dataset, respectively.
    \item \textbf{Raw}. To test the effectiveness of pre-training, we directly trained the full Transformer on the ED dataset, and then compared it with the fine-tuned models.
\end{itemize}
Note that we did not include the EmoPrepend-1 model by~\citet{DBLP:conf/acl/RashkinSLB19} as our baseline, because in their paper, its human evaluation performance is actually reported to be worse than the fine-tuned Transformer. All the models have a hidden size of 300, and were trained until the minimum validation loss was reached. For inference we used beam search with beam size 32 and 4-gram repeats blocking. Further details regarding the implementation parameters can be found in Appendix~\ref{appendix:implement}.

\subsection{Automatic Evaluation}
\label{subsec:auto_eval}

\begin{table*}[t]
    \small
    \centering
    \begin{tabular}{l rrrr rrrr rrrr}
        \toprule
        & \multicolumn{4}{c}{\textbf{OS}} & \multicolumn{4}{c}{\textbf{EDOS}} & \multicolumn{4}{c}{\textbf{ED}} \\
        \cmidrule(lr){2-5} \cmidrule(lr){6-9} \cmidrule(lr){10-13}
        \textbf{Model} & \textbf{PPL} & \textbf{D1} & \textbf{D2} & \textbf{SES} & \textbf{PPL} & \textbf{D1} & \textbf{D2} & \textbf{SES} & \textbf{PPL} & \textbf{D1} & \textbf{D2} & \textbf{SES} \\
        \midrule
        Pre-trained (OS) & 24.8 & .046 & .159 & .172 & 37.8 & .046 & .154 & .126 & 564.6 & .044 & .167 & .178 \\
        Fine-tuned (EDOS) & 26.9 & .044 & .139 & .162 & 32.3 & .056 & .165 & .137 & 452.6 & .031 & .107 & .176 \\
        Fine-tuned (ED) & 88.9 & .030 & .109 & \textbf{.174} & 140.8 & .028 & .096 & .130 & 19.3 & .026 & .091 & \textbf{.316} \\
        Raw (ED) & 793.9 & .009 & .032 & .144 & 1615.0 & .008 & .027 & .098 & 35.8 & .008 & .029 & .278 \\
        \midrule
        MEED2 (OS) & \textbf{22.0} & \textbf{.064} & \textbf{.210} & .168 & 31.9 & .061 & .197 & .130 & 487.3 & .046 & .171 & .174 \\
        MEED2 (OS$\,\to\,$EDOS) & 22.8 & .057 & .196 & .168 & \textbf{28.5} & \textbf{.070} & \textbf{.225} & \textbf{.171} & 391.7 & \textbf{.051} & \textbf{.199} & .207 \\
        MEED2 (OS$\,\to\,$ED) & 84.3 & .038 & .153 & .165 & 125.7 & .036 & .138 & .116 & \textbf{17.2} & .036 & .140 & .299 \\
        \bottomrule
    \end{tabular}
    \caption{Automatic evaluation results. Here PPL denotes perplexity, D1 and D2 denote Distinct-1 and -2, and SES denotes the sentence embedding similarity. X$\,\to\,$Y means pre-training on X and then fine-tuning on Y.}
    \label{tab:auto_eval}
\end{table*}

\begin{table*}[t]
    \small
    \centering
    \begin{tabular}{l rrr rrr rrr}
        \toprule
        & \multicolumn{3}{c}{\textbf{OS}} & \multicolumn{3}{c}{\textbf{EDOS}} & \multicolumn{3}{c}{\textbf{ED}} \\
        \cmidrule(lr){2-4} \cmidrule(lr){5-7} \cmidrule(lr){8-10}
        \textbf{Model} & \textbf{P} & \textbf{R} & \textbf{F-1} & \textbf{P} & \textbf{R} & \textbf{F-1} & \textbf{P} & \textbf{R} & \textbf{F-1} \\
        \midrule
        Random & .1484 & .0240 & .0285 & .0382 & .0250 & .0266 & .0989 & .0165 & .0215 \\
        MEED2 (OS) & \textbf{.2210} & \textbf{.3960} & .2312 & .0109 & .1040 & .0198 & .0942 & .3070 & .1442 \\
        MEED2 (OS$\,\to\,$EDOS) & .2012 & .1480 & .1537 & \textbf{.1029} & \textbf{.1495} & \textbf{.0917} & .1288 & .2630 & .1674 \\
        MEED2 (OS$\,\to\,$ED) & .2166 & .3265 & \textbf{.2502} & .0253 & .0870 & .0239 & \textbf{.2660} & \textbf{.3530} & \textbf{.2864} \\
        \bottomrule
    \end{tabular}
    \caption{Weighted precision, recall and F-1 scores of the response emotion/intent predictor in our model on the three datasets. X$\,\to\,$Y means pre-training on X and then fine-tuning on Y.}
    \label{tab:emo_pred_scores}
\end{table*}

Most of the existing automatic metrics directly compare the generated response with the ground-truth provided by human, often in a simple way. Due to the inherent diversity of human conversations, this is not suitable for dialog models, since for the same prompt, there could exist many responses that are equally good. In fact, \citet{DBLP:conf/emnlp/LiuLSNCP16} has shown that word-overlap-based metrics (specifically BLEU~\cite{DBLP:conf/acl/PapineniRWZ02}, METEOR~\cite{DBLP:conf/acl/BanerjeeL05}, and ROUGE~\cite{lin-2004-rouge}) and word embedding metrics all exhibit weak or no correlation with human judgements. To this end, we did not adopt these metrics in our experiment, but instead considered the following:
\begin{itemize}
    \item \textbf{Perplexity}. Perplexity is a model-dependent metric that measures how well a probability model predicts a given sample. In our case, a lower perplexity score indicates better capability of generating the ground-truth response.
    \item \textbf{Distinct-1 and -2}. The Distinct-1 and -2 metrics~\cite{DBLP:conf/naacl/LiGBGD16} measure the diversity of the generated responses by calculating the ratio of unique unigrams or bigrams over the total number of unigrams or bigrams in the generated responses.
    \item \textbf{Sentence Embedding Similarity}. For this metric, we use Sentence-BERT~\cite{DBLP:conf/emnlp/ReimersG19} to obtain an embedding for the generated response as well as the ground-truth, and then calculate the cosine similarity between the two embeddings.
\end{itemize}

The results of automatic evaluation are shown in Table~\ref{tab:auto_eval}. Our model (MEED2) achieves lower perplexity scores than the corresponding full Transformer on all the three datasets. Here we have an extra model configuration, Raw (ED), to compare with Fine-tuned (ED), in order to see the effects brought by pre-training. As we can see, without pre-training on OS, the model gets much worse performance on the perplexity scores. This indicates that pre-training and then fine-tuning is preferred to directly training on a target dataset. On Distinct-1 \mbox{and -2}, our model always has a higher score than the corresponding full Transformer model, suggesting that by injecting additional emotion information, the dialog system could be guided to generate more diverse responses. We also observe that on the ED dataset, our model fine-tuned on EDOS actually has the highest Distinct scores, even though it has never seen the ED data. We conjecture that this is because the EDOS dataset is much bigger than the ED dataset, and contains text that is more diverse. Table~\ref{tab:emo_pred_scores} lists the weighted precision, recall, and F-1 scores of the response emotion/intent predictor for different model configurations.

If we consider a zero-shot setting, meaning the model is evaluated on data from a different domain than its training data, we see from Table~\ref{tab:auto_eval} that all models achieves higher perplexity scores on zero-shot test data. In particular, models trained on the OS (EDOS) dataset achieves lower perplexity on the EDOS (OS) dataset, compared with the results on the ED dataset. This is because OS and EDOS dialogs are actually curated from the same source, while the source of ED data is quite different. Moreover, models trained on EDOS has better perplexity scores on OS dataset, due to the performance boost brought by fine-tuning.

\subsection{Human Evaluation via Crowdsourcing}

\begin{table*}[t]
    \small
    \centering
    \begin{tabular}{l rrr rrr rrr}
        \toprule
        & \multicolumn{3}{c}{\textbf{OS}} & \multicolumn{3}{c}{\textbf{EDOS}} & \multicolumn{3}{c}{\textbf{ED}} \\
        \cmidrule(lr){2-4} \cmidrule(lr){5-7} \cmidrule(lr){8-10}
        \textbf{Model} & \textbf{Good} & \textbf{Okay} & \textbf{Bad} & \textbf{Good} & \textbf{Okay} & \textbf{Bad} & \textbf{Good} & \textbf{Okay} & \textbf{Bad} \\
        \midrule
        Pre-trained (OS) & .3097 & .2878 & .4025 & .2975 & .2933 & .4091 & .1799 & .3037 & .5164 \\
        MEED2 (OS) & .3166 & .3158 & .3676 & .3073 & .3288 & .3639 & .1863 & .3088 & .5049 \\
        MEED2 (OS$\,\to\,$EDOS) & .3175 & .3036 & .3789 & .2926 & .3034 & .4040 & .2097 & .2891 & .5012 \\
        MEED2 (OS$\,\to\,$ED) & \textbf{.3513} & .3125 & .3362 & \textbf{.3535} & .3093 & .3372 & \textbf{.4890} & .3033 & .2077 \\
        \bottomrule
    \end{tabular}
    \caption{Human evaluation results on each of the three test sets. Numbers have been normalized across the three quality categories on each test set. X$\,\to\,$Y means pre-training on X and then fine-tuning on Y.}
    \label{tab:human_eval_ind}
\end{table*}

\begin{figure*}
    \centering
    \includegraphics[width=\textwidth]{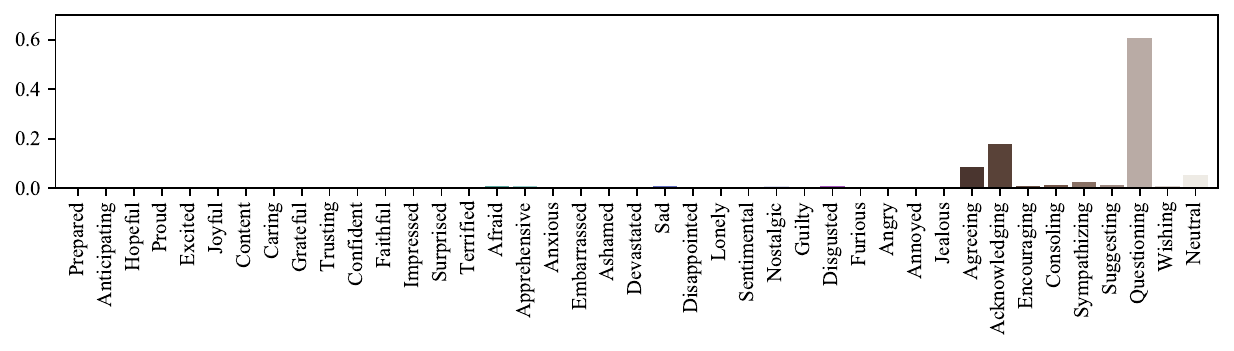}
    \caption{Distribution of emotions/intents in the responses generated by MEED2 (OS$\,\to\,$ED) rated as \emph{good}.}
    \label{fig:pred_emot_dist}
\end{figure*}

\begin{table*}[t]
    \small
    \centering
    \begin{tabular}{p{15.5cm}}
        \toprule
        \textbf{Context (from OS)}: \\
        - Well, now, you know, most of your average people... won 't know a venomous reptile from a large garden worm. \\
        \textbf{Pre-trained (OS)}: I don 't know what you're talking about. \\
        \textbf{MEED2 (OS) (questioning)}: What do you mean, you won 't know? \\
        \textbf{MEED2 (OS$\,\to\,$EDOS) (terrified)}: I'm not afraid of snakes. \\
        \textbf{MEED2 (OS$\,\to\,$ED) (terrified)}: Oh my gosh, I hate snakes. What kind of snake was it? \\
        \textbf{Ground-truth (afraid)}: Oh, no. It has to be an absolute, authentic, poisonous snake... or he won 't work with it. \\
        \midrule
        \textbf{Context (from EDOS)}: \\
        - Yes, it's been a long wait. The second half of King of Comedy will now begin! \\
        - Hyo-jin. Yes. I'm excited to see what stars the second half will bring. \\
        - Yes. Welcome our first contestants. Who are they? \\
        \textbf{Pre-trained (OS)}: The second half will bring the second half will be the second half. \\
        \textbf{MEED2 (OS) (questioning)}: Who are they? What do they look like? \\
        \textbf{MEED2 (OS$\,\to\,$EDOS) (excited)}: Let's begin! \\
        \textbf{MEED2 (OS$\,\to\,$ED) (excited)}: I can't wait to see who they are. \\
        \textbf{Ground-truth (wishing)}: Warriors. Welcome! \\
        \midrule
        \textbf{Context (from ED)}: \\
        - Even though I could've gotten more money at another job, I've stayed with my current employer. It is a good place to work. \\
        - well that's good to hear I guess. what do you do? \\
        - I work in public relations. I create campaigns for companies. \\
        \textbf{Pre-trained (OS)}: what do you do? \\
        \textbf{MEED2 (OS) (questioning)}: what do they do in public relations? \\
        \textbf{MEED2 (OS$\,\to\,$EDOS) (questioning)}: What are you doing in public affairs? \\
        \textbf{MEED2 (OS$\,\to\,$ED) (acknowledging)}: that sounds like a great job. \\
        \textbf{Ground-truth (questioning)}: what's your most successful campaign so far? \\
        \bottomrule
    \end{tabular}
    \caption{Some samples of the generated responses. X$\,\to\,$Y means pre-training on X and then fine-tuning on Y.}
    \label{tab:sample}
\end{table*}

Human evaluation for dialog models has been widely adopted due to the limitations of automatic metrics. However, the experiment should be carefully designed so that the raters clearly understand the instructions and are constantly engaged in the evaluation tasks. Moreover, most of the existing work only recruited a limited number of raters to evaluate a test set of small size, therefore leading to possibly biased results. In this paper, we carefully designed a human evaluation experiment that enables the raters to work on the evaluation tasks more easily and at the same time keeps them engaged by incorporating bonus checkpoints.

\subsubsection{A New Evaluation Strategy}
We conducted our human evaluation experiment on Amazon Mechanical Turk (MTurk). The 6,000 test dialogs were randomly shuffled and then split into 600 Human Intellligence Tasks (HITs), with each HIT containing 10 dialogs to be evaluated. For each test dialog, we included the generated responses from four candidate models, i.e., Pre-trained (OS), MEED2 (OS), MEED2 (OS$\,\to\,$EDOS), and MEED2 (OS$\,\to\,$ED). Existing human experiments in dialog evaluation adopt either Likert Scale or side-by-side comparison (A/B testing). Likert Scale allows accurate evaluation of single items, but lacks reference and comparison; while A/B testing allows comparison, it doesn't scale. Our method is the first one that combines these two strategies and leverages on the merits of both. We allow the workers to drag and drop multiple candidate responses to one of the three pre-defined areas: \emph{good}, \emph{okay}, and \emph{bad}, according to whether the response is emotionally appropriate following the given dialog context. In this way, it is easier for the workers to finish the tasks, and we also benefit from the accurate scoring results. In order to make the workers more engaged in the evaluation, and also encourage those providing high-quality answers, for each HIT we attached a bonus task to three ED dialogs, by adding the ground-truth response as a candidate. If the worker successfully put the ground-truth into the \emph{good} or \emph{okay} category, he or she will receive a bonus point. We gave a bonus of \$0.1 to those workers who obtained all the three bonus points. More details of the human evaluation setup, including screenshots of the interface, can be found in Appendix~\ref{appendix:human_eval_setup}.

\subsubsection{Human Evaluation Results}
In total we received 24,000 answers from the MTurk experiment (4 answers for each of the 6,000 dialogs). We discarded answers from low-quality workers, i.e., those who provided the same answer for almost all dialogs, and those who completed the tasks in less than five minutes and failed to obtain at least two bonus points. Then, to calculate the human evaluation scores, we further selected out those assignments with at least two bonus points, and obtained a total number of 21,630 answers. The human evaluation results on the three individual test sets are shown in Table~\ref{tab:human_eval_ind}. From the table we see that our model outperforms the full Transformer on all three datasets (Pre-trained (OS) v.s.~MEED2 (OS)), and of all the four model configurations, our model trained on ED achieves the highest percentage of good response on all three datasets, meaning training on ED enables the model to gain both good held-out performance and good zero-shot performance. Compared with our model only pre-trained on OS, it achieves better performance on OS and ED if fine-tuned on EDOS, but not on EDOS itself, meaning this model has a good zero-shot performance but the held-out performance is somehow lower. This could be explained by the unbalanced emotion/intent distribution in the OS dataset. As discussed in Section~\ref{subsec:auto_eval}, for our model trained on OS, the response emotion/intent predictor would usually predict the dominating ``questioning'' category. For EDOS dialogs, since the response emotion/intent is more difficult to predict, responding in questions is probably safer.

We also investigated the distribution of emotions/intents in the generated responses, to see which emotions/intents are more preferred by the workers. For responses generated by MEED2 (OS$\,\to\,$ED) that are rated as \emph{good}, we gathered the predicted emotions/intents and calculated a probability distribution over the 41 categories, which is shown in Figure~\ref{fig:pred_emot_dist}. We can see that \emph{questioning}, \emph{acknowledging} and \emph{agreeing} are the major categories. This shows that our model tends to generate responses with the empathetic intents, and they are indeed more preferred by the human evaluators.

\subsection{Case Study}

In this section, we give some sample responses generated by the models in Table~\ref{tab:sample}. We took one dialog from each test set (OS, EDOS and ED). We can observe that most of the generated responses are syntactically correct (exceptional cases are from Pre-trained (OS)). The models could understand the dialog context and generate appropriate responses. For example, in the first dialog, our models fine-tuned on EDOS and ED recognize and understand the word ``reptile'' in the context, and then as response, generate the word ``snakes''. We can also observe from the table that the response emotions predicted by our models (fine-tuned on EDOS and ED) are reasonable and follow the emotions embedded in the dialog context. Moreover, the generated responses are indeed consistent with the predicted emotions. Note that our model trained on OS has a big chance of predicting the ``questioning'' category, which is due to the unbalanced distribution in the training set. More samples of the generated responses can be found in Appendix~\ref{appendix:more_samples}.

\section{Conclusion}
In this paper, we emphasize the importance of incorporating more fine-grained empathetic response intents into the design of empathetic dialog models. To this end, we proposed an empathetic dialog model capable of learning the emotion/intent interactions from the dialog data at a more precise level, and producing empathetic responses accordingly. To facilitate the training process, we also curated a large-scale dialog dataset from the OpenSubtitles corpus. Pre-training dialog models on this dataset could largely boost the performance of down-stream empathetic response generation. Our model was evaluated through a carefully designed human evaluation experiment on the crowdsourcing platform, on a large test set never attempted before. As future work, we would like to improve the accuracy of the response emotion/intent predictor in the model, which we found plays a vital role in generating empathetic responses.

\bibliography{custom}
\bibliographystyle{acl_natbib}

\appendix
\section{The Cleaning Procedure of the OpenSubtitles Dialogs}
After segmenting the subtitle files in the OpenSubtitles corpus into dialogs, we further clean the dataset with the following steps:
\label{appendix:cleaning}
\begin{itemize}
    \item Remove redundant spaces in the utterances (e.g., spaces at the beginning and the end, and unnecessary spaces between the tokens);
    \item Remove utterances starting with ``previously on \dots'' (narration at the beginning of TV episodes);
    \item Remove utterances that simply repeat previous turns;
    \item Remove utterances that do not start with alphabet, digit, ``\texttt{\textquotesingle}''~(single quote), or ``\texttt{"}''~(double quote);
    \item For utterances in the form of ``character : \dots'', remove the character information and keep the remaining part;
    \item Remove utterances with length (number of tokens) less than 2 or greater than 100;
    \item Remove utterances with percentage of alphabet letters less than 60\%;
    \item Remove utterances with percentage of distinct tokens less than $2/3$;
    \item Reduce frequency of any utterance to 100.
\end{itemize}
Whenever we remove an utterance, we discard all the following utterances in the same dialog.

\section{Implementation Parameters}
\label{appendix:implement}
\begin{table*}[t]
    \centering
    \begin{tabular}{lrrrr}
        \toprule
        \textbf{Model} & \textbf{\# Parameters} & \textbf{\# Training Epochs} & \textbf{Training Time} & \textbf{Validation PPL} \\
        \midrule
        Pre-trained (OS) & 121M & 50 epochs & 171.00 hr & 24.51 \\
        Fine-tuned (EDOS) & 121M & 5 epochs & 4.23 hr & 31.78 \\
        Fine-tuned (ED) & 121M & 9 epochs & 19.50 min & 21.04 \\
        Raw (ED) & 121M & 55 epochs & 1.87 hr & 40.56 \\
        \midrule
        MEED2 (OS) & 180M & 50 epochs & 181.38 hr & 21.70 \\
        MEED2 (OS$\,\to\,$EDOS) & 180M & 6 epochs & 4.88 hr & 28.12 \\
        MEED2 (OS$\,\to\,$ED) & 180M & 10 epochs & 20.09 min & 19.02 \\
        \bottomrule
    \end{tabular}
    \caption{Training details and validation performance of each model configuration.}
    \label{tab:num_epochs}
\end{table*}

Here we summarize some of the parameters of the model implementation:
\begin{itemize}
    \item We use the RoBERTa tokenizer to tokenize the input utterances, and the vocabulary size is 50,265. We allow a maximum number of 100 tokens as the input to the model.
    \item We use 4 sub-layers in the encoder and decoder, with 6 heads in the multi-head attention. The dimension of the hidden units is 300, and the dimension of the pointwise feed-forward layers is 1200. We use a dropout rate of 0.1, and the GELU~\cite{hendrycks2016gaussian} activation function for the hidden layers.
    \item The loss function is optimized with the Adam optimizer~\cite{DBLP:journals/corr/KingmaB14} with an initial learning rate of $5\times 10^{-5}$.
    \item For inference, we use beam search with a beam size of 32. To prevent the models from generating repetitive tokens or $n$-grams, we modified the beam search algorithm so that at each time step, if any of the branches contains repetitive 4-grams, we set the log probability of this branch to infinitely negative, to stop it from being further expanded.
\end{itemize}
All the models were trained with a batch size of 512, on machines with 4 Nvidia Titan X Pascal GPUs, 2 Intel Xeon E5-2680 v3 CPUs, and 256GB RAM. Table~\ref{tab:num_epochs} lists the training details as well as the validation performance for all the models.

\section{Human Evaluation Setup}
\label{appendix:human_eval_setup}
The 6,000 test dialogs were split into 600 HITs, with each HIT containing 10 dialogs to be evaluated. We allowed a maximum of 4 workers working on the same HIT, and gave \$0.4 for completing a HIT. When launching the experiment, we only included workers from English speaking countries, i.e., US, AU, NZ, GB, and CA. We also required the workers to have at least 100 approved assignments, and the approval rate is at least 95\%. To avoid having the same worker working on too many HITs, we ran a custom script at the backend that constantly checked the worker statistics and blocked the worker if he/she had already finished 50 HITs.

\begin{figure}[t]
    \centering
    \includegraphics[width=\columnwidth]{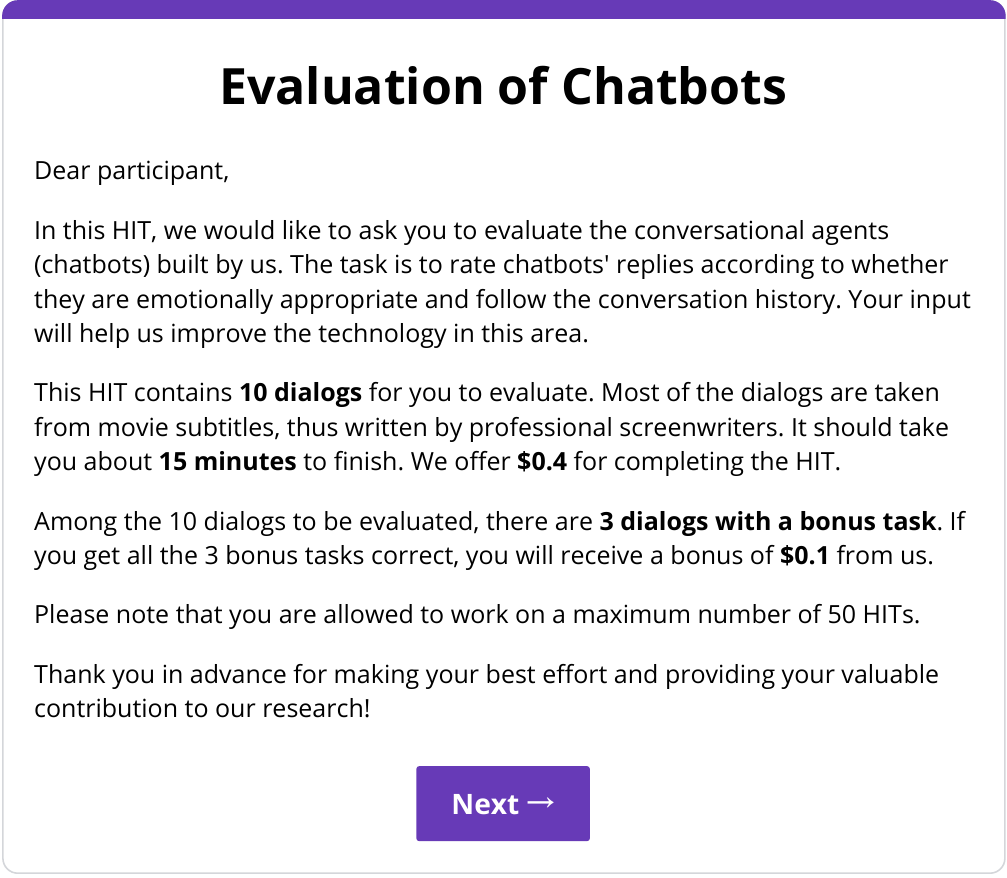}
    \caption{A screenshot of the welcome page of our human evaluation experiment.}
    \label{fig:human_eval_welcome}
\end{figure}

\begin{figure*}[t]
    \centering
    \begin{subfigure}[t]{0.475\textwidth}
        \includegraphics[width=\columnwidth]{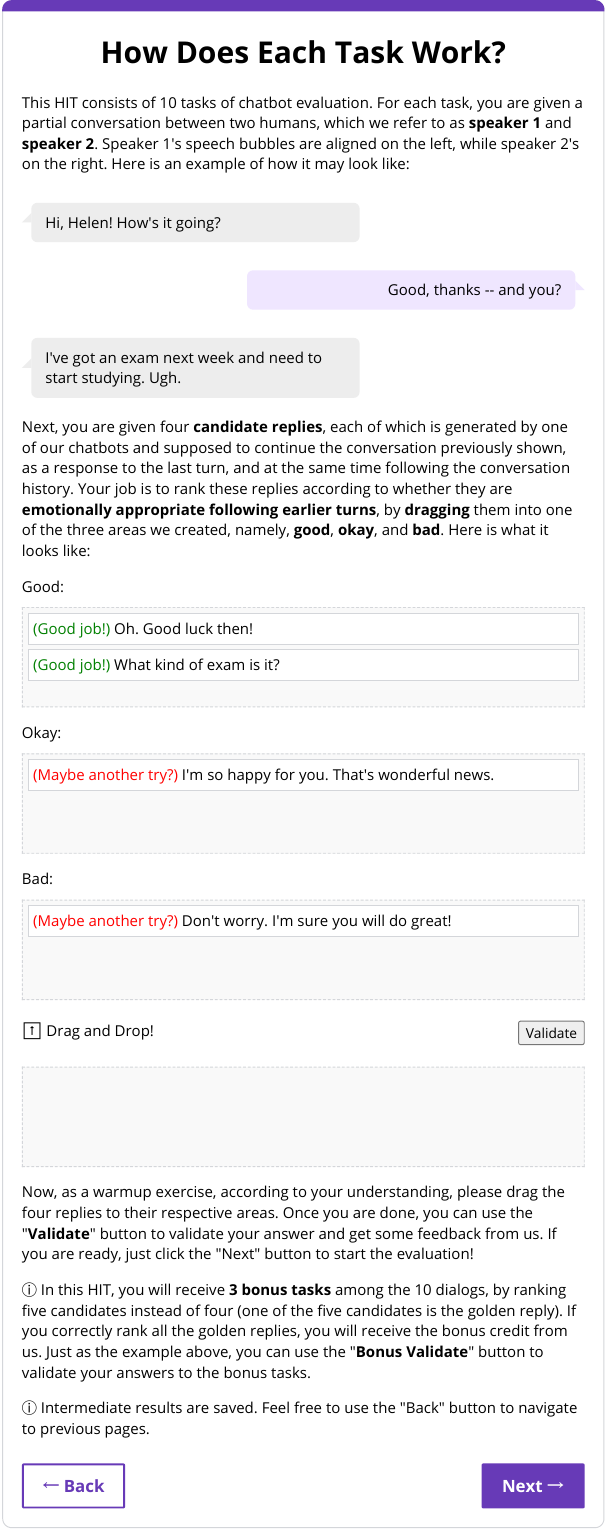}
        \caption{A screenshot of the instruction page of our human evaluation experiment.}
        \label{fig:human_eval_instruct}
    \end{subfigure}
    \quad
    \begin{subfigure}[t]{0.475\textwidth}
        \includegraphics[width=\columnwidth]{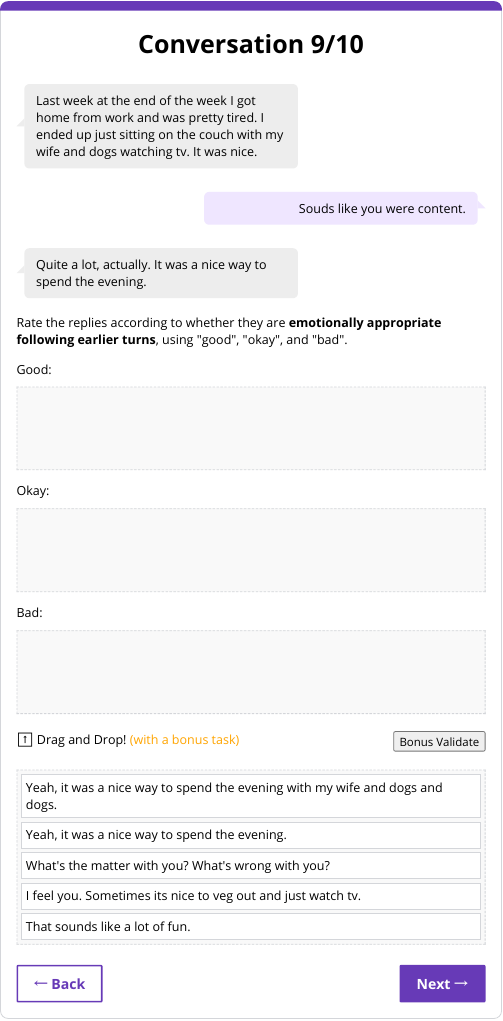}
        \caption{A screenshot of the task page of our human evaluation experiment. This task includes a bonus checkpoint.}
        \label{fig:human_eval_task}
    \end{subfigure}
    \caption{Screenshots of our human evaluation experiment.}
    \label{fig:human_eval_exp}
\end{figure*}

Figure~\ref{fig:human_eval_welcome} is a screenshot of the welcome page of our human evaluation experiment on the crowdsourcing platform. Figure~\ref{fig:human_eval_instruct} shows the instructions and explains to the worker how the tasks work, where the worker can also try an example task by dragging and dropping the candidate responses to one of the defined areas, and then validate the answer and get the feedback. Figure~\ref{fig:human_eval_task} is a screenshot of the task page. This task includes a bonus checkpoint, meaning one of the candidate responses is the ground-truth. The worker can click the ``Bonus Validation'' button to check if he/she has successfully obtained the bonus point.

\section{More Samples of Model Outputs}
\label{appendix:more_samples}
Table~\ref{tab:more_sample} lists some more samples of the generated responses, with dialog contexts taken from the OS, EDOS, and ED datasets.

\begin{table*}[t]
    \small
    \centering
    \begin{tabular}{p{15.5cm}}
        \toprule
        \textbf{Context (OS)}: \\
        - What do you think happened to the lights, huh? \\
        \textbf{Pre-trained (OS)}: I don 't know. I've never seen anything like it. \\
        \textbf{MEED2 (OS) (questioning)}: What do you think happened to the lights? \\
        \textbf{MEED2 (OS$\,\to\,$EDOS) (afraid)}: I don 't know. I can 't see anything. \\
        \textbf{MEED2 (OS$\,\to\,$ED) (afraid)}: I don't know what happened to the lights. \\
        \textbf{Ground-truth (agreeing)}: Our waiter told me that the power in Bogota is not very reliable. That's a lie. You and I both know it. They cut the lights on purpose! I know how you feel. And I want her back just as much as you do. \\
        \midrule
        \textbf{Context (OS)}: \\
        - Oh, this is really your day. \\
        - Having trouble with your little Brunhilde, dear? \\
        - All right then, tell me about your lady wrestler friend. \\
        \textbf{Pre-trained (OS)}: I don 't know what you're talking about. \\
        \textbf{MEED2 (OS) (questioning)}: What's she like? \\
        \textbf{MEED2 (OS$\,\to\,$EDOS) (jealous)}: I don 't know what she's talking about. I've never seen her before. \\
        \textbf{MEED2 (OS$\,\to\,$ED) (questioning)}: What did she do to you? \\
        \textbf{Ground-truth (questioning)}: Who is she? An old flame, darling? \\
        \midrule
        \textbf{Context (EDOS)}: \\
        - I'm telling you. The black kid's fixed the launch pad and we're outta here. Future's looking dim. Soon they won 't have money for soda. Let's go party somewhere else. \\
        - Party? Don 't you understand? If God had wanted us to have fun instead of the Ten Commandments he would have sent the Ten Suggestions. \\
        - Man, what have they done to you? \\
        \textbf{Pre-trained (OS)}: What have they done to me? \\
        \textbf{MEED2 (OS) (questioning)}: What have they done to you? \\
        \textbf{MEED2 (OS$\,\to\,$EDOS) (faithful)}: We're not going anywhere. We're staying here. \\
        \textbf{MEED2 (OS$\,\to\,$ED) (neutral)}: I don't know what they have done to me, but I'm glad it worked out for you. \\
        \textbf{Ground-truth (content)}: Go in peace. I'm staying put. \\
        \midrule
        \textbf{Context (EDOS)}: \\
        - Sinbad, what are you doing? It's just a bull up ahead. \\
        - Come on! I already took one of your eyes! I'm your monster, you're going! \\
        - Come get me! Come on! \\
        \textbf{Pre-trained (OS)}: I don 't know what you're talking about. \\
        \textbf{MEED2 (OS) (questioning)}: Sinbad, what are you doing? \\
        \textbf{MEED2 (OS$\,\to\,$EDOS) (angry)}: Come on, Sinbad! \\
        \textbf{MEED2 (OS$\,\to\,$ED) (angry)}: I hate when that happens. \\
        \textbf{Ground-truth (furious)}: No! The last! \\
        \midrule
        \textbf{Context (ED)}: \\
        - Sometimes an old song runs through my head and I remember the times where I used to listen to songs so many times that I memorized the words \\
        - That sounds nice. It's always good to remember better times. What kind of music? \\
        - Like old kind of emo songs, full of angst and whatnot \\
        \textbf{Pre-trained (OS)}: Whatnot? \\
        \textbf{MEED2 (OS) (questioning)}: Don 't you think that's funny? \\
        \textbf{MEED2 (OS$\,\to\,$EDOS) (nostalgic)}: Like old kind of music \\
        \textbf{MEED2 (OS$\,\to\,$ED) (nostalgic)}: Those are the good old days. \\
        \textbf{Ground-truth (nostalgic)}: Ha, I used to listen to that stuff back in high school. The My Chemical Romance gets me every time. \\
        \midrule
        \textbf{Context (ED)}: \\
        - I recently went to eat a restaurant that was not very clean. The whole place had an odd odor and made me uncomfortable. \\
        \textbf{Pre-trained (OS)}: What are you doing here? Why are you here? \\
        \textbf{MEED2 (OS) (questioning)}: What kind of smell? \\
        \textbf{MEED2 (OS$\,\to\,$EDOS) (questioning)}: What's that smell? \\
        \textbf{MEED2 (OS$\,\to\,$ED) (disgusted)}: That's gross. What kind of smell was it? \\
        \textbf{Ground-truth (acknowledging)}: Ew, that is not good! I wouldn't have been able to stay and eat. \\
        \bottomrule
    \end{tabular}
    \caption{More samples of the generated responses with dialog contexts taken from the OS, EDOS, and ED datasets.}
    \label{tab:more_sample}
\end{table*}

\end{document}